\documentclass[conference]{IEEEtran}
\IEEEoverridecommandlockouts

\newif\ifblind 
\blindfalse   

\usepackage{gensymb}
\usepackage{float}
\usepackage{multirow}
\usepackage{booktabs}
\usepackage{subcaption}
\usepackage{cite}
\usepackage{amsmath,amssymb,amsfonts}
\usepackage{algorithmic}
\usepackage{graphicx}
\usepackage{textcomp}
\usepackage{xcolor}
\usepackage{amsmath}

\def\BibTeX{{\rm B\kern-.05em{\sc i\kern-.025em b}\kern-.08em
    T\kern-.1667em\lower.7ex\hbox{E}\kern-.125emX}}
\begin{document}

\title{Inspector: Conversational and Lightweight Analyzer of Analog Circuit Layouts Using LLM and CNNs
\thanks{This work was supported by Italian Ministero dell'Universita e della Ricerca (MUR)'s Fondo Italiano Scienze Applicate (FISA) under Grant no. FISA-2024-00333.}
}

\ifblind
    \author{
    \IEEEauthorblockN{Anonymous Authors}
    }
\else
    \author{\IEEEauthorblockN{1\textsuperscript{st} Abril Cano Castro}
    \IEEEauthorblockA{\textit{DEIB} \\
    \textit{Politecnico di Milano}\\
    Milano, Italy \\
    abril.cano@mail.polimi.it}
    \and
    \IEEEauthorblockN{2\textsuperscript{nd} Giuseppe Chiari}
    \IEEEauthorblockA{\textit{DEIB} \\
    \textit{Politecnico di Milano}\\
    Milano, Italy \\
    giuseppe.chiari@polimi.it}
    \and
    \IEEEauthorblockN{3\textsuperscript{rd} Michele Piccoli}
    \IEEEauthorblockA{\textit{DEIB} \\
   	\textit{Politecnico di Milano}\\
   	Milano, Italy \\
    michele.piccoli@polimi.it}
    \and
    \IEEEauthorblockN{4\textsuperscript{th} Federico Viola}
    \IEEEauthorblockA{\textit{DEIB} \\
   	\textit{Politecnico di Milano}\\
   	Milano, Italy \\
    federico.viola@polimi.it}
    \and
    \IEEEauthorblockN{5\textsuperscript{th} Davide Zoni}
    \IEEEauthorblockA{\textit{DEIB} \\
   	\textit{Politecnico di Milano}\\
   	Milano, Italy \\
    davide.zoni@polimi.it}
    }
\fi

\maketitle

\begin{abstract}
The integration of artificial intelligence into computer-aided design frameworks has sparked a shift in the design of analog integrated circuits~(ICs), transitioning the field from using manual and algorithmic-based solutions to adopting automated and intelligent paradigms. In this scenario, the GDSII file represents the industry-standard database containing the ultimate and most accurate source of information of the analog circuit, encapsulating the complex physical geometries and parasitic realities that define tape out performance. This paper proposes a novel framework that combines fine-tuned LLMs and CNNs to analyze GDSII files of analog circuits, enabling a conversational interface between the tool and the designers. Experimental results using thousands of analog designs across four realistic tasks demonstrate that the proposed solution outperforms state-of-the-art general-purpose massive VLMs by a significant margin~(up to $81\%$), thus providing a lightweight solution to the problem of GDSII analysis.
\end{abstract}


\section{Introduction}
\label{sec:introduction}
%
Analog design remains largely manual, heuristic-driven, and time-consuming, especially in the back-end phase, where GDSII~(Graphic Data System II) layouts are used for physical implementation, parasitic extraction, and final verification. Although artificial intelligence~(AI) has advanced analog computer-aided design~(CAD), most progress targets front-end tasks, while GDSII analysis still depends on manual inspection or rigid, computationally expensive rule-based scripts~\cite{KMS+2019,XZL+2019}. This limits the extraction of high-level semantic information, such as device counts or sub-circuit topologies, from increasingly complex layouts.
Building on AI assistants for software engineering and digital RTL (Register Transfer Level) design~\cite{BGK+2023,CWR+2023}, we introduce a lightweight semantic approach that enables natural-language queries over physical layouts, shifting layout analysis from passive verification toward active conversational intelligence.
This paper proposes \emph{Inspector}, a framework based on fine-tuned LLMs~(Large Language Models) and CNNs~(Convolutional Neural Networks). CNNs have been widely adopted as detectors across diverse domains, including construction~\cite{LFH+2021}, cybersecurity~\cite{CGL+2024,GCZ2024,GCZ2025}, and autonomous driving~\cite{CLL+2021}.
\newline
In \emph{Inspector}, CNNs enable accurate recognition of GDSII images, while the LLM handles user interaction and interprets tasks and results. This hybrid design preserves vision-language interaction while remaining orders of magnitude smaller than VLMs~(Vision Language Models).
\begin{figure*}[t]
	\centering
	\begin{subfigure}{\textwidth}
		\centering
		\includegraphics[width=0.9\linewidth]{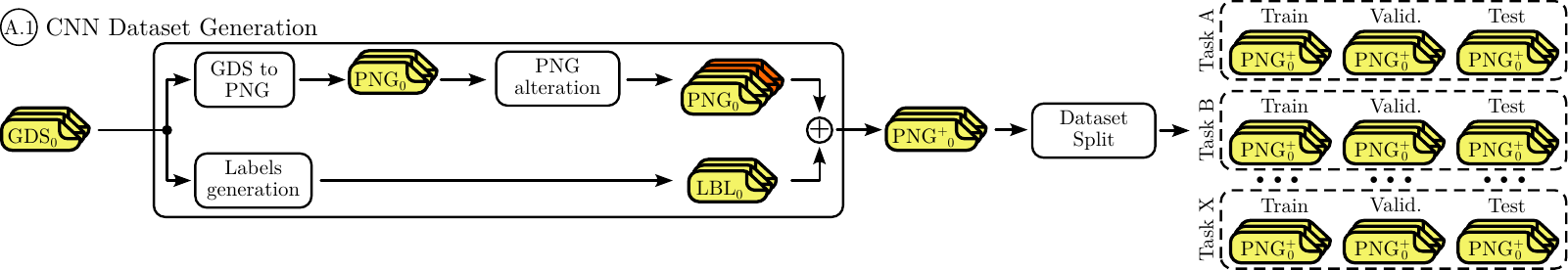}
		\caption{Dataset generation workflow for CNN training. 
			GDS layouts are converted into PNG images, augmented through controlled alterations, automatically labeled~(LBL), and partitioned into task-specific training, validation, and test subsets.}
		\label{sfig:cnn_dataset_flow}
	\end{subfigure}
	\begin{subfigure}{\textwidth}
		\centering
		\includegraphics[width=0.8\linewidth]{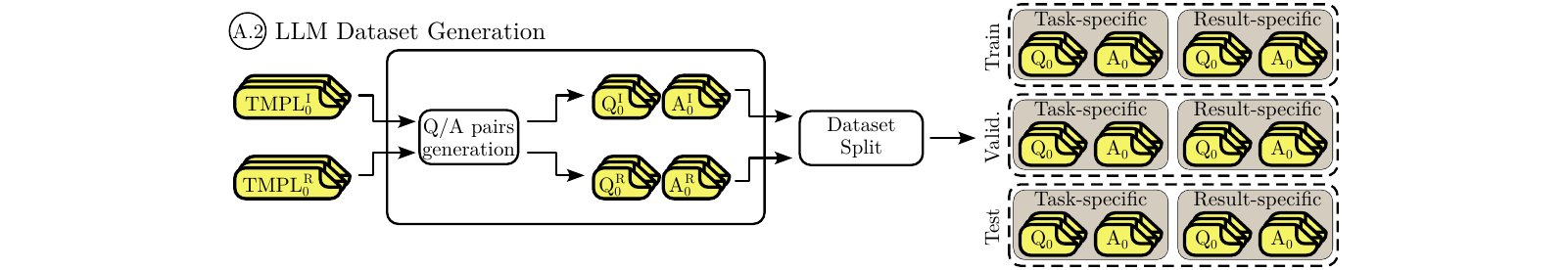}
		\caption{Dataset generation workflow for LLM training. 
			Q/A pairs are derived from templates for \textit{Task Identification} and \textit{Result Reconstruction}, then partitioned into training, validation, and test subsets.}
		\label{sfig:llm_dataset_flow}
	\end{subfigure}
	\caption{Datasets generation flows: CNN Dataset Generation~(top) and LLM Dataset Generation~(bottom).}
	\label{fig:dataset_generation}
\end{figure*}
\emph{Inspector} is evaluated on four realistic design tasks over diverse layouts, outperforming state-of-the-art general-purpose VLMs by up to $81\%$ in task accuracy while using a smaller model~($\sim$1B parameters).
In particular, this work delivers three contributions to the state of the art:
\begin{itemize}
	\item Lightweight conversational model for GDSII analysis, fine-tuned for analog-layout understanding and natural-language interaction.
	\item Open dataset and end-to-end workflow\footnote{https://github.com/hardware-fab/Inspector}, including layout images, question-answer conversations, and a reproducible training/evaluation pipeline.
	\item Experimental validation on real layouts, against state-of-the-art VLM baselines, using representative DRC-free and LVS-compliant analog circuits.
\end{itemize}
%

	

\begin{figure*}[t]
	\includegraphics[width=1\linewidth]{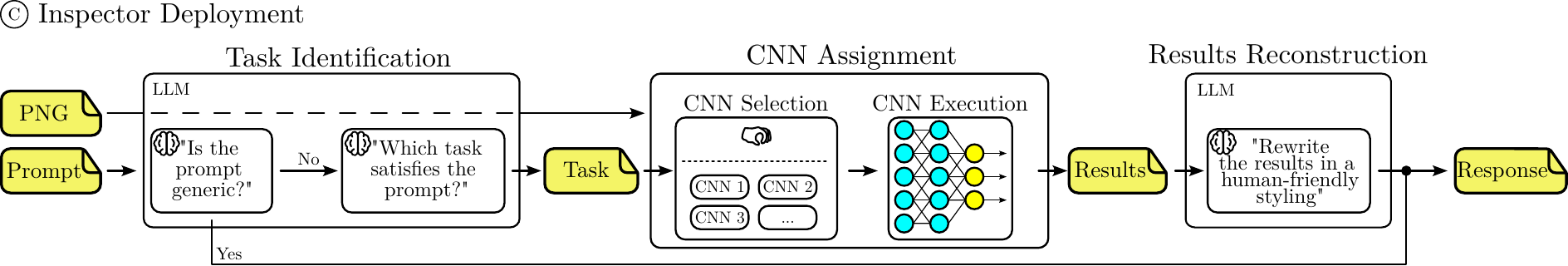}
	\caption{Overview of the \emph{Inspector Deployment} phase, where the LLM identifies the task, selects the corresponding CNN, and reconstructs the CNN output as a natural-language response.}
	\label{fig:inspector_deployment}
\end{figure*}

\section{State of the art}
\label{sec:state_of_the_art}

\noindent\textbf{Front-end --} AI methods have been extensively studied for schematic analysis, topology generation, and circuit optimization. Netlistify~\cite{HCH+2025} combines CNNs and transformers to recover components, orientations, and connections from analog schematics, while LaMAGIC~\cite{CYS+2024}, AnalogGenie~\cite{GCY2025}, and AnalogCoder~\cite{YSG+2024} leverage LLM- and code-based representations for topology synthesis. LLMs have also enabled autonomous design agents: Artisan~\cite{CHL+2024} adopts tree-of-thought and chain-of-thought prompting for operational-amplifier design, AnalogXpert~\cite{HSY+2025} exploits subcircuit libraries with iterative refinement, and ADO-LLM~\cite{YWX+2024} and LEDRO~\cite{DHA+2025} integrate optimization workflows.

\noindent\textbf{Back-end --} Physical-layout analysis remains less explored, and direct semantic understanding of GDSII is still limited. Recent work includes SOLOMON~\cite{BX2025}, which generates layout scripts through tree-of-thought reasoning, LLM-HD~\cite{CWW+2024} for lithography-hotspot detection from GDSII binaries, and DRC-Coder~\cite{CHL+2025}, translating textual design rules into verification code. 
\newline
In analog layout automation, BAG~\cite{CPL+2013,CHB+2018}, ALIGN~\cite{KMS+2019}, and MAGICAL~\cite{XZL+2019} provide generator- and algorithm-driven flows, complemented by neural~\cite{BBL+2021}, Bayesian~\cite{BZC+2023,GZY+2024}, reinforcement-learning~\cite{BBV+2024,BBV+2025a,DBB+2025,CPZ+2026}, and graph-neural-network approaches~\cite{BBV+2025a}.

\section{Methodology}
\label{sec:methodology}

\subsection{Dataset Generation}
\label{ssec:dataset_gen}
Figure~\ref{fig:dataset_generation} illustrates the data-generation workflows used for \emph{Inspector} calibration, namely CNN training and LLM fine-tuning.
As shown in Figure~\ref{sfig:cnn_dataset_flow}, CNN data are produced from GDSII layouts by converting each design into a rasterized PNG image, applying controlled graphical alterations, and automatically extracting labels~(LBL) with component classes and bounding boxes. Images and labels are combined into paired image-annotation samples~(PNG$^+$), then partitioned into task-specific training, validation, and test subsets for visual detection.
Figure~\ref{sfig:llm_dataset_flow} reports the LLM dataset flow. Predefined textual templates emulate user interactions for two objectives: \emph{(i)} task identification, where the LLM selects the requested operation, and \emph{(ii)} result reconstruction, where structured CNN outputs are translated into natural-language responses. The resulting Q/A pairs are split into training, validation, and test sets for supervised fine-tuning.

\begin{table*}[t]
	\centering
	\caption{Dataset composition. For each category, the table reports the circuit types, the number of layout variants, the average number of devices per variant, and the average per-device-type counts~(NMOS, PMOS, capacitors, and resistors).}
	\label{tbl:dataset_composition}
	\resizebox{1\linewidth}{!}{%
		\begin{tabular}{llccccccc}
			\toprule
			\textbf{Category} & \textbf{Circuit Type} & \textbf{Variants} & \textbf{Avg. Devices} & \textbf{NMOS} & \textbf{PMOS} & \textbf{CAP} & \textbf{RES} & \textbf{Total Devices} \\
			\midrule
			\multirow{5}{*}{\shortstack{Single\\Component}} & \quad Capacitor & $5\,000$ & $1$ & -- & -- & $1$ & -- & $5\,000$ \\
			& \quad NMOS Transistor & $5\,000$ & $1$ & $1$ & -- & -- & -- & $5\,000$ \\
			& \quad PMOS Transistor & $5\,000$ & $1$ & -- & $1$ & -- & -- & $5\,000$ \\
			& \quad Resistor & $5\,000$ & $1$ & -- & -- & -- & $1$ & $5\,000$ \\
			\cmidrule(lr){1-9}
			& \quad Subtotal & $20\,000$ & -- & -- & -- & -- & -- & $20\,000$ \\
			\midrule
			\multirow{7}{*}{\shortstack{Base\\Circuits}} & \quad Ahuja OTA & $995$ & $15$ & $10$ & $4$ & $1$ & -- & $14\,925$ \\
			& \quad Gate Driver & $1\,000$ & $10$ & $4$ & $4$ & -- & $2$ & $10\,000$ \\
			& \quad High-Pass Filter (HPF) & $962$ & $13$ & $5$ & $3$ & $3$ & $2$ & $12\,506$ \\
			& \quad Low-Dropout Regulator (LDO) & $989$ & $9$ & $3$ & $3$ & $1$ & $2$ & $8\,901$ \\
			& \quad Low-Pass Filter (LPF) & $971$ & $13$ & $5$ & $3$ & $3$ & $2$ & $12\,623$ \\
			& \quad Miller OTA & $977$ & $13$ & $5$ & $4$ & $2$ & $2$ & $12\,701$ \\
			\cmidrule(lr){1-9}
			& \quad Subtotal & $5\,894$ & -- & -- & -- & -- & -- & $71\,656$ \\
			\midrule
			\multirow{1}{*}{\shortstack{Mixed}} & \quad Mixed Topologies & $4\,140$ & $21.6$ & $9.7$ & $6.8$ & $2.3$ & $2.8$ & $89\,424$ \\
			\midrule
			& \textbf{Total Dataset} & \textbf{$30\,034$} & -- & -- & -- & -- & -- & \textbf{$181\,080$} \\
			\bottomrule
		\end{tabular}
	}
\end{table*}

	

%
\subsection{Inspector Training}
\label{ssec:pipeline_training}
The training phase is divided into two parallel workflows to address the distinct learning objectives of the different models.
The \textit{LLM Fine-tuning} stage trains a pre-trained language model using the Q/A pairs generated for both the \textit{Task Identification} and \textit{Result Reconstruction} steps, allowing the model to specialize its semantic understanding and reasoning capabilities for the target domain.
In contrast, \textit{CNN Training} focuses on visual learning: depending on the desired task, the appropriate set of augmented \texttt{PNG$^{+}$} samples is selected and used to train a dedicated convolutional neural network. This separation enables each model to exploit the data representation best suited to its specific objective while contributing to the overall framework.
\begin{table}[t]
	\centering
	\caption{Task descriptions, organized by complexity \emph{(i)}~Easy, \emph{(ii)}~Medium, and \emph{(iii)}~Hard.}
	\label{tbl:training_tasks}
	\footnotesize
	\resizebox{1\linewidth}{!}{%
		\begin{tabular}{lllr}
			\toprule
			\textbf{Complexity} & \textbf{Task ID} & \textbf{Task Description}  \\
			\midrule
			\multirow{2}{*}{\textbf{Easy}} 
			& \multirow{2}{*}{\shortstack{A}} & Identification of single component devices \\
			&   & (capacitors, resistors, NMOS, PMOS)  \\
			\midrule
			\multirow{4}{*}{\textbf{Medium}} 
			& \multirow{2}{*}{\shortstack{B}} & Identification of base circuit topologies \\
			&   & (OTAs, filters, regulators, gate drivers) \\
			& \multirow{2}{*}{\shortstack{C}} & Component counting and enumeration \\
			&   & in base circuits &  \\
			\midrule
			\multirow{2}{*}{\textbf{Hard}} 
			& \multirow{2}{*}{\shortstack{D}} & Component counting and enumeration \\
			&   & in complex mixed circuits  \\
			\bottomrule
		\end{tabular}
	}
\end{table}
\subsection{Inspector Deployment}
\label{ssec:pipeline_deployment}
Figure~\ref{fig:inspector_deployment} illustrates the deployment phase of \emph{Inspector}, where a user prompt and a PNG representation of the GDSII layout are transformed into a human-readable response.
\newline The pipeline comprises three stages: \emph{(i) Task Identification}, \emph{(ii) CNN Assignment}, and \emph{(iii) Results Reconstruction}. First, the LLM classifies the prompt as either generic or task-specific; generic queries are answered directly, while task-specific queries produce a task label. 
\newline This label selects the corresponding CNN, which analyzes the layout image and returns structured detections, including component classes and spatial information. Finally, the LLM interprets these detections in the context of the original prompt and generates the final natural-language answer.

\section{Experimental Evaluation}
\label{sec:experimental_evaluation}
\subsection{Experimental Setup}
\label{ssec:exp_setup}
Table~\ref{tbl:dataset_composition} summarizes the CNN training dataset, which includes single-device layouts, basic analog building blocks, and mixed topologies obtained from structured combinations of the base circuits. All layouts are implemented in the SkyWater 130nm PDK and satisfy both DRC and LVS requirements.
We evaluate \emph{Inspector} on the four tasks defined in Section~\ref{ssub:tasks_definition} and compare it with two representative VLM baselines: \emph{InternLM-4KHD}~\cite{XPY+2024}, a high-resolution multimodal model, and \emph{GPT-5.2}~\cite{gpt}, used as a large-scale commercial reference.
\begin{table*}[t]
\centering
\small
\caption{Comparison of \emph{Inspector} with state-of-the-art VLMs in terms of parameters, runtime, and accuracy.}
\resizebox{0.9\linewidth}{!}{%
    \begin{tabular}{c|rrrr|rrr|rrr}
    \toprule
    \textbf{Task}
    & \multicolumn{4}{c|}{\textbf{Inspector}}
    & \multicolumn{3}{c|}{\textbf{InternLM-4KHD~\cite{XPY+2024}}}
    & \multicolumn{3}{c}{\textbf{GPT-5.2~\cite{gpt}}} \\
    
    \cmidrule(lr){2-5}\cmidrule(lr){6-8}\cmidrule(lr){9-11}
    
    & parameters       & $t_{t+f}$    & $t_{i}$ & accuracy
    & parameters       & $t_{i}$              & accuracy    
    & parameters       & $t_{i}$              & accuracy     \\

    & (\# in billions) & (mm:ss)    & (ss)     & (\%)   
    & (\# in billions) & (ss)                  & (\%)  
    & (\# in billions) & (ss)                  & (\%)         \\

    \midrule
    \textbf{A}  
    & $\sim1$          & $29$:$25$  & $1.19$    & $97$
    & $7$              & $0.52$     & $41$ 
    & $50\,000^\ast$   & $2.3$      & $100$ \\
    
    \textbf{B} 
    & $\sim1$          & $28$:$48$  & $0.24$  & $91$ 
    & $7$              & $0.54$     & $16$
    & $50\,000^\ast$   & $2.8$      & $83$ \\
    
    \textbf{C} 
    & $\sim1$          & $19$:$04$    & $3.65$ & $99$
    & $7$              & $0.54$       & $18$
    & $50\,000^\ast$   & $4$          & $20$ \\
    
    \textbf{D} 
    & $\sim1$          & $15$:$37$    & $2.62$ & $92$
    & $7$              & $0.52$       & $24$
    & $50\,000^\ast$   & $3.8$        & $26$ \\

    \bottomrule
    \end{tabular}%
}
	
	\begin{minipage}{0.95\linewidth}
        \vspace{1mm}
		\footnotesize
		($\ast$)~The parameter count for GPT-5.2 is an estimate, as the exact model size has not been publicly disclosed.
	\end{minipage}

\label{tbl:results}
\end{table*}
\begin{figure}[t]
	\centering
	\includegraphics[width=0.7\linewidth]{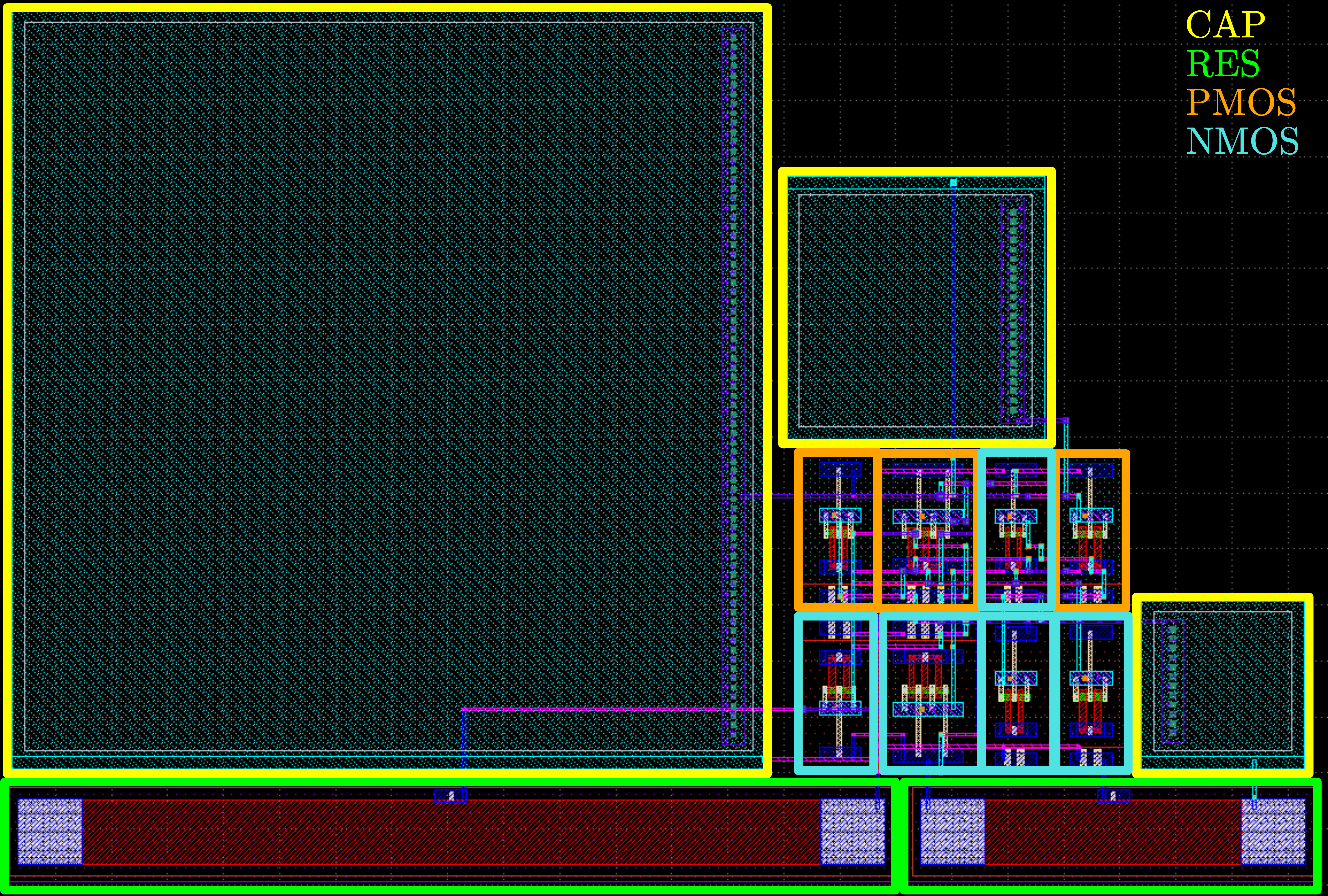}
	\caption{Layout of an HPF. The labels are generated to contain the whole bounding box of the device.}
	\label{fig:layout_labeled}
\end{figure}
\begin{figure*}[t]
	\centering
	    
    \begin{subfigure}[t]{0.22\textwidth}
    	\centering
    	\includegraphics[width=\linewidth]{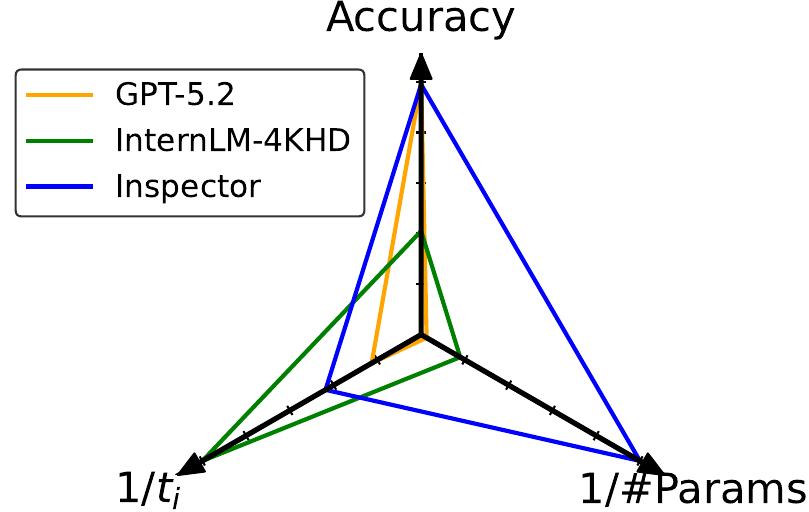}
        \caption{Task A.}
    	\label{sfig:exp_res_task_a}
    \end{subfigure}
    \hfill
    \begin{subfigure}[t]{0.22\textwidth}
    	\centering
    	\includegraphics[width=\linewidth]{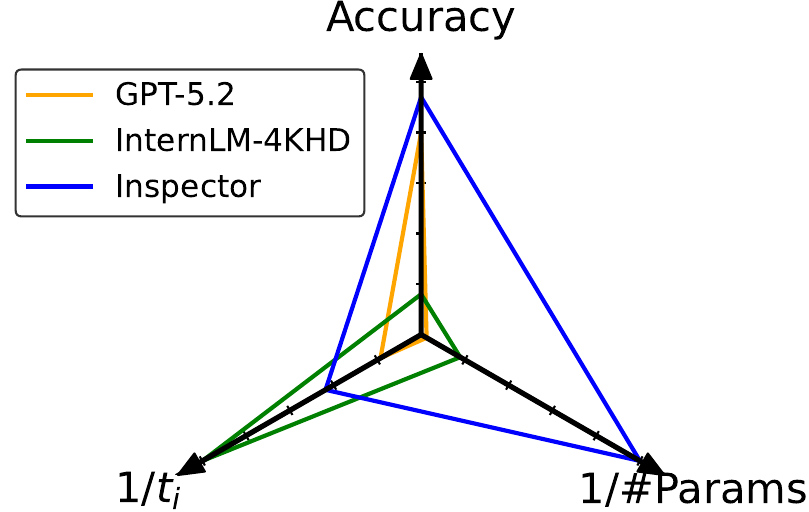}
        \caption{Task B.}
    	\label{sfig:exp_res_task_b}
    \end{subfigure}
    \hfill
    \begin{subfigure}[t]{0.22\textwidth}
    	\centering
    	\includegraphics[width=\linewidth]{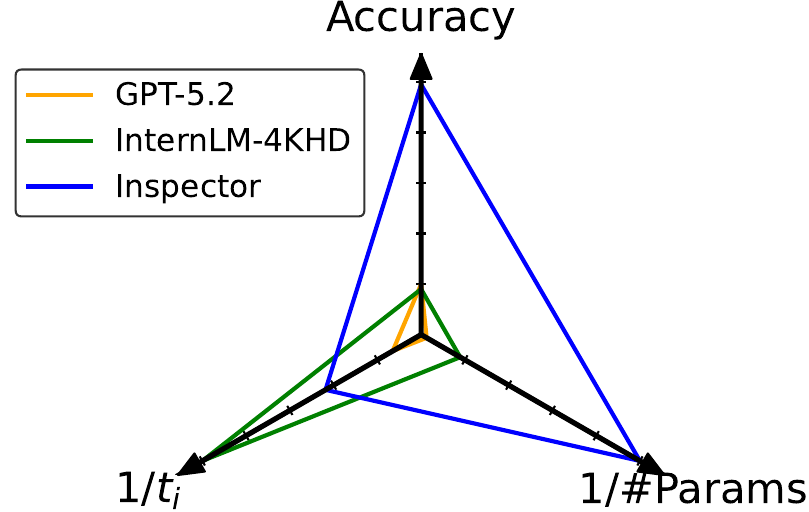}
        \caption{Task C.}
    	\label{sfig:exp_res_task_c}
    \end{subfigure}
    \hfill
    \begin{subfigure}[t]{0.22\textwidth}
    	\centering
    	\includegraphics[width=\linewidth]{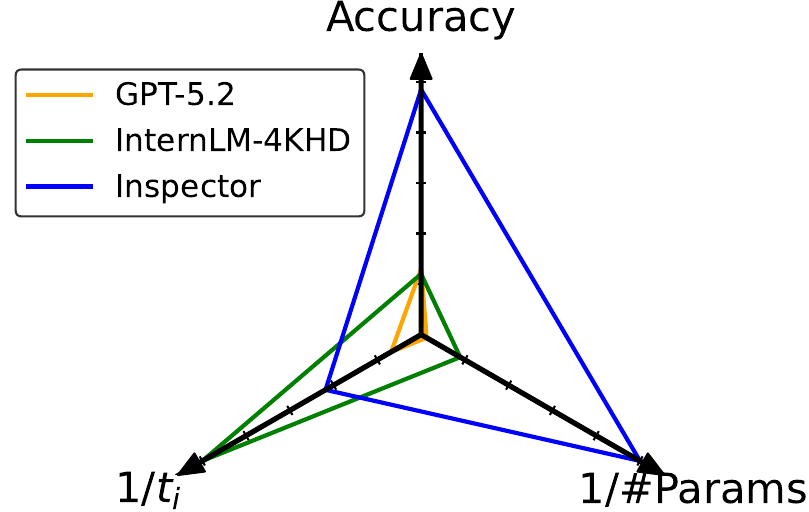}
        \caption{Task D.}
    	\label{sfig:exp_res_task_d}
    \end{subfigure}
    
	\caption{Triangular trade-off among accuracy, execution time, and parameter count. Lower-is-better metrics are shown as reciprocals.}
	\label{fig:exp_res}
\end{figure*}
\subsection{Tasks Definition}
\label{ssub:tasks_definition}
Table~\ref{tbl:training_tasks} groups the evaluated tasks by complexity. Task~A identifies individual devices; Tasks~B and~C respectively recognize base circuit topologies and enumerate their components.
\newline Finally, Task~D extends component counting to mixed circuits composed of multiple interconnected blocks. Figure~\ref{fig:layout_labeled} shows an enriched PNG$^{+}$ example generated from a GDSII high-pass filter layout with automatically produced component labels.

\subsection{Metrics Definition}
\emph{Inspector} is evaluated through both CNN- and LLM-oriented metrics in order to assess the performance of its visual recognition and semantic reasoning components. For the CNN stage, we report mAP@0.5 and mAP@0.5:0.95 to measure detection accuracy across different Intersection-over-Union thresholds, together with precision~($TP/(TP+FP)$) and recall~($TP/(TP+FN)$), which quantify the trade-off between false positives and false negatives. For the LLM stage, we report Task Identification~(T-ID), defined as the percentage of prompts correctly routed to the appropriate analysis pipeline, and Results Reconstruction~(R-R), which measures the percentage of natural-language responses correctly reconstructed from the structured outputs produced by the CNN models.
\subsection{Experimental Results}
\label{ssec:exp_res}
Table~\ref{tbl:results} compares \emph{Inspector} with standalone VLMs. Its accuracy combines mAP@0.5, T-ID, and R-R from Table~\ref{tbl:cnn_detailed_metrics} and Table~\ref{tbl:llm_detailed_metrics}; runtimes and parameter counts aggregate the CNN and LLM stages.
\emph{Inspector} reaches $91$--$99$\% accuracy with second-scale inference using Llama-3.2~\cite{GDJ+2024} and lightweight task-specific YOLOv8~\cite{yolov8} detectors. InternLM-4KHD remains below $41$\%, while GPT-5.2 drops from perfect Task~A accuracy to $26$\% on Task~D. Figure~\ref{fig:exp_res} summarizes the resulting trade-off.
\begin{table}[t]
\centering
\caption{\emph{Inspector}'s CNN detection performance.}
\label{tbl:cnn_detailed_metrics}
\resizebox{\linewidth}{!}{%
	\begin{tabular}{c|rrrr}
		\toprule
		\textbf{Task} & \textbf{mAP@0.5 } & \textbf{mAP@0.5:0.95 } & \textbf{Precision } & \textbf{Recall } \\
		& \textbf{(\%)} & \textbf{(\%)} & \textbf{(\%)} & \textbf{(\%)} \\
		\midrule
		\textbf{A} & $99.4$ & $99.48$ & $99.91$ & $100$ \\
		\textbf{B} & $99.3$ & $99.34$ & $98.43$ & $98.12$ \\
		\textbf{C} & $99.5$ & $98.66$ & $99.99$ & $99.99$ \\
		\textbf{D} & $99.4$ & $93.13$ & $99.55$ & $99.40$ \\
		\bottomrule
	\end{tabular}
}
\end{table}
Table~\ref{tbl:cnn_detailed_metrics} shows mAP@0.5 above 99\% for all tasks, with near-perfect precision and recall.
\begin{table}[t]
	\centering
	\small
	\caption{\emph{Inspector}'s LLM performance metrics.}
	\label{tbl:llm_detailed_metrics}
	\resizebox{0.9\columnwidth}{!}{%
		\begin{tabular}{c|cc|ccc}
			\toprule
			\textbf{Task} 
			& \multicolumn{2}{c|}{\textbf{Accuracy (\%)}} 
			& \textbf{Params} 
			& \textbf{Fine-t. Time} 
			& \textbf{Inf. Time} \\
			\cmidrule(lr){2-3}
			& \emph{T-ID} & \emph{R-R} 
			& (\#B) & (s) & (mm:ss) \\
			\midrule
			\textbf{A} & $97.95$ & $100$   & $1$ & $17$.$32$ & $01:18$ \\ 
			\textbf{B} & $93.68$ & $98.41$ & $1$ & $14$.$46$ & $00:19$ \\ 
			\textbf{C} & $100$   & $99.58$ & $1$ & $15$.$04$ & $04:03$ \\ 
			\textbf{D} & $100$   & $92.98$ & $1$ & $14$.$40$ & $02:59$ \\ 
			\bottomrule
		\end{tabular}
	}
\end{table}
Table~\ref{tbl:llm_detailed_metrics} shows T-ID and R-R between approximately $92$\% and $100$\%, with second-scale inference.


\begin{figure}[t]
	\centering
	\includegraphics[width=0.7\linewidth]{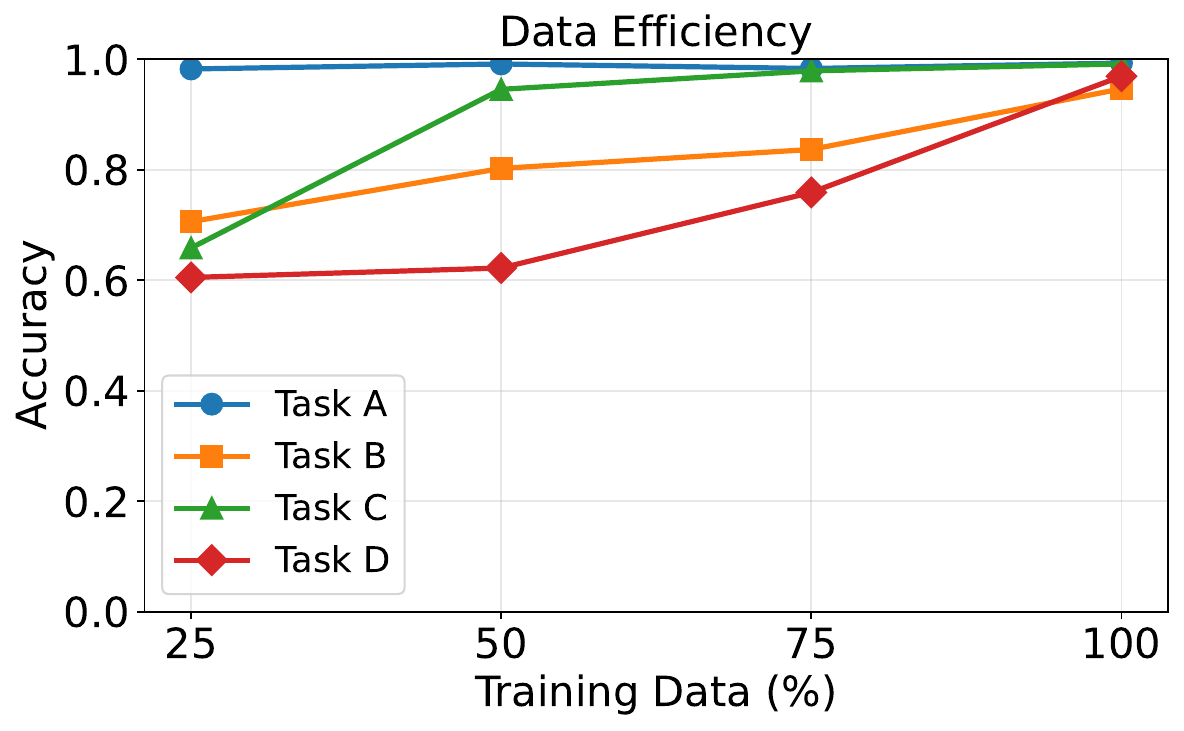}
    \caption{Accuracy ablation with respect to training data percentage, showing task sensitivity to data availability and pipeline scalability.}
	\label{fig:ablation_study}
\end{figure}

\subsection{Ablation Study}
\label{ssec:ablation_study}
The ablation study in Figure~\ref{fig:ablation_study} evaluates the effect of training data availability on task accuracy. 
\newline While Task~A reaches near-optimal performance with only $50\%$ of the data, indicating limited supervision requirements, Tasks~B--D exhibit a more progressive improvement as the dataset grows. In particular, Task~D shows the highest sensitivity to data reduction, reflecting the greater structural complexity and variability associated with higher-level reasoning tasks.

\section{Conclusions}
\label{sec:conclusions}
This paper presents \emph{Inspector}, a hybrid LLM-CNN framework for conversational analysis of analog circuit layouts directly from GDSII data. 
By combining semantic reasoning with specialized visual detection, the proposed pipeline enables designers to query layout information through a natural language interface.
Experimental results on multiple tasks and thousands of designs show that \emph{Inspector} significantly outperforms state-of-the-art general-purpose VLMs, achieving up to $81\%$ higher accuracy while maintaining a lightweight architecture. 

\bibliographystyle{IEEEtran}
\bibliography{iclad_2026}

@misc{gpt,
	title={GPT-5.2}, 
	author={OpenAI},
	year={2025},
	url={https://openai.com/index/introducing-gpt-5-2/}, 
}

@misc{yolov8,
	title={YOLOv8},
	author={Jocher, Glenn and Qiu, Jing and Chaurasia, Ayush},
	year={2023},
	url={https://github.com/ultralytics/ultralytics},
}

@article{GDJ+2024,
	title={The llama 3 herd of models},
	author={Grattafiori, Aaron and Dubey, Abhimanyu and Jauhri, Abhinav and Pandey, Abhinav and Kadian, Abhishek and Al-Dahle, Ahmad and Letman, Aiesha and Mathur, Akhil and Schelten, Alan and Vaughan, Alex and others},
	journal={arXiv preprint arXiv:2407.21783},
	year={2024}
}

@misc{XPY+2024,
      title={InternLM-XComposer2-4KHD: A Pioneering Large Vision-Language Model Handling Resolutions from 336 Pixels to 4K HD}, 
      author={Xiaoyi Dong and Pan Zhang and Yuhang Zang and Yuhang Cao and Bin Wang and Linke Ouyang and Songyang Zhang and Haodong Duan and Wenwei Zhang and Yining Li and Hang Yan and Yang Gao and Zhe Chen and Xinyue Zhang and Wei Li and Jingwen Li and Wenhai Wang and Kai Chen and Conghui He and Xingcheng Zhang and Jifeng Dai and Yu Qiao and Dahua Lin and Jiaqi Wang},
      year={2024},
      eprint={2404.06512},
      archivePrefix={arXiv},
      primaryClass={cs.CV},
      url={https://arxiv.org/abs/2404.06512}, 
}

@INPROCEEDINGS{HCH+2025,
  author={Huang, Chun-Yen and Chen, Hsuan-I and Ho, Hao-Wen and Kang, Pei-Hsin and Lin, Mark Po-Hung and Liu, Wen-Hao and Ren, Haoxing},
  booktitle={2025 ACM/IEEE 7th Symposium on Machine Learning for CAD (MLCAD)}, 
  title={Netlistify: Transforming Circuit Schematics into Netlists with Deep Learning}, 
  year={2025},
  volume={},
  number={},
  pages={1-8},
  doi={10.1109/MLCAD65511.2025.11189145}}

@inproceedings{CWW+2024,
	author = {Chen, Yuyang and Wu, Yiwen and Wang, Jingya and Wu, Tao and He, Xuming and Yu, Jingyi and Geng, Hao},
	title = {LLM-HD: Layout Language Model for Hotspot Detection with GDS Semantic Encoding},
	year = {2024},
	isbn = {9798400706011},
	publisher = {Association for Computing Machinery},
	address = {New York, NY, USA},
	url = {https://doi.org/10.1145/3649329.3658479},
	doi = {10.1145/3649329.3658479},
	booktitle = {Proceedings of the 61st ACM/IEEE Design Automation Conference},
	articleno = {121},
	numpages = {6},
	location = {San Francisco, CA, USA},
	series = {DAC '24}
}

@inproceedings{CHL+2025, 
   series={ISPD ’25},
   title={DRC-Coder: Automated DRC Checker Code Generation Using LLM Autonomous Agent},
   url={http://dx.doi.org/10.1145/3698364.3705347},
   DOI={10.1145/3698364.3705347},
   booktitle={Proceedings of the 2025 International Symposium on Physical Design},
   publisher={ACM},
   author={Chang, Chen-Chia and Ho, Chia-Tung and Li, Yaguang and Chen, Yiran and Ren, Haoxing},
   year={2025},
   month=mar, pages={143–151},
   collection={ISPD ’25} 
}

@inproceedings{BX2025,
	title={Enhancing Reasoning to Adapt Large Language Models for Domain-Specific Applications},
	author={Bo Wen and Xin Zhang},
	booktitle={Adaptive Foundation Models: Evolving AI for Personalized and Efficient Learning},
	year={2024},
	url={https://openreview.net/forum?id=F8rniHIK3H}
}

@misc{YSG+2024,
      title={AnalogCoder: Analog Circuit Design via Training-Free Code Generation}, 
      author={Yao Lai and Sungyoung Lee and Guojin Chen and Souradip Poddar and Mengkang Hu and David Z. Pan and Ping Luo},
      year={2024},
      eprint={2405.14918},
      archivePrefix={arXiv},
      primaryClass={cs.LG},
      url={https://arxiv.org/abs/2405.14918}, 
}

@inproceedings{YWX+2024, 
   series={ICCAD ’24},
   title={ADO-LLM: Analog Design Bayesian Optimization with In-Context Learning of Large Language Models},
   url={http://dx.doi.org/10.1145/3676536.3676816},
   DOI={10.1145/3676536.3676816},
   booktitle={Proceedings of the 43rd IEEE/ACM International Conference on Computer-Aided Design},
   publisher={ACM},
   author={Yin, Yuxuan and Wang, Yu and Xu, Boxun and Li, Peng},
   year={2024},
   month=oct, pages={1–9},
   collection={ICCAD ’24} }

@inproceedings{CYS+2024,
	title={LaMAGIC: Language-Model-based Topology Generation for Analog Integrated Circuits},
	author={Chang, Chen-Chia and Shen, Yikang and Fan, Shaoze and Li, Jing and Zhang, Shun and Cao, Ningyuan and Chen, Yiran and Zhang, Xin},
	booktitle={Forty-first International Conference on Machine Learning},
	year={2024},
}

@inproceedings{CHL+2024,
	author = {Chen, Zihao and Huang, Jiangli and Liu, Yiting and Yang, Fan and Shang, Li and Zhou, Dian and Zeng, Xuan},
	title = {Artisan: Automated Operational Amplifier Design via Domain-specific Large Language Model},
	year = {2024},
	isbn = {9798400706011},
	publisher = {Association for Computing Machinery},
	address = {New York, NY, USA},
	url = {https://doi.org/10.1145/3649329.3655903},
	doi = {10.1145/3649329.3655903},
	booktitle = {Proceedings of the 61st ACM/IEEE Design Automation Conference},
	articleno = {39},
	numpages = {6},
	location = {San Francisco, CA, USA},
	series = {DAC '24}
}

@misc{DHA+2025,
	title={Ledro: Llm-enhanced design space reduction and optimization for analog circuits},
	author={Kochar, Dimple Vijay and Wang, Hanrui and Chandrakasan, Anantha P and Zhang, Xin},
	booktitle={2025 IEEE International Conference on LLM-Aided Design (ICLAD)},
	pages={141--148},
	year={2025},
	organization={IEEE}
}

@misc{HSY+2025,
	title={Analogxpert: Automating analog topology synthesis by incorporating circuit design expertise into large language models},
	author={Zhang, Haoyi and Sun, Shizhao and Lin, Yibo and Wang, Runsheng and Bian, Jiang},
	booktitle={2025 International Symposium of Electronics Design Automation (ISEDA)},
	pages={772--777},
	year={2025},
	organization={IEEE}
}

@inproceedings{CPL+2013,
	title={BAG: A designer-oriented integrated framework for the development of AMS circuit generators},
	author={Crossley, John and Puggelli, Alberto and Le, H-P and Yang, B and Nancollas, R and Jung, Kwangmo and Kong, Lingkai and Narevsky, Nathan and Lu, Yue and Sutardja, Nicholas and others},
	booktitle={2013 IEEE/ACM International Conference on Computer-Aided Design (ICCAD)},
	year={2013},
	organization={IEEE}
}

@inproceedings{CHB+2018,
	title={BAG2: A process-portable framework for generator-based AMS circuit design},
	author={Chang, Eric and Han, Jaeduk and Bae, Woorham and Wang, Zhongkai and Narevsky, Nathan and Nikolic, Borivoje and Alon, Elad},
	booktitle={2018 IEEE Custom Integrated Circuits Conference (CICC)},
	year={2018},
}

@inproceedings{KMS+2019,
	title={ALIGN: Open-source analog layout automation from the ground up},
	author={Kunal, Kishor and Madhusudan, Meghna and Sharma, Arvind K and Xu, Wenbin and Burns, Steven M and Harjani, Ramesh and Hu, Jiang and Kirkpatrick, Desmond A and Sapatnekar, Sachin S},
	booktitle={Proceedings of the 56th Annual Design Automation Conference 2019},
	year={2019}
}

@inproceedings{XZL+2019,
	title={MAGICAL: Toward fully automated analog IC layout leveraging human and machine intelligence},
	author={Xu, Biying and Zhu, Keren and Liu, Mingjie and Lin, Yibo and Li, Shaolan and Tang, Xiyuan and Sun, Nan and Pan, David Z},
	booktitle={2019 IEEE/ACM International Conference on Computer-Aided Design (ICCAD)},
	year={2019},
	organization={IEEE}
}

@inproceedings{BBL+2021,
	title={Dnn-opt: An rl inspired optimization for analog circuit sizing using deep neural networks},
	author={Budak, Ahmet F and Bhansali, Prateek and Liu, Bo and Sun, Nan and Pan, David Z and Kashyap, Chandramouli V},
	booktitle={2021 58th ACM/IEEE Design Automation Conference (DAC)},
	pages={1219--1224},
	year={2021},
	organization={IEEE}
}

@article{GZY+2024,
	title={Post-layout simulation driven analog circuit sizing},
	author={Gao, Xiaohan and Zhang, Haoyi and Ye, Siyuan and Liu, Mingjie and Pan, David Z and Shen, Linxiao and Wang, Runsheng and Lin, Yibo and Huang, Ru},
	journal={Science China Information Sciences},
	volume={67},
	number={4},
	pages={142401},
	year={2024},
	publisher={Springer}
}

@inproceedings{BZC+2023,
	title={Joint optimization of sizing and layout for ams designs: Challenges and opportunities},
	author={Budak, Ahmet F and Zhu, Keren and Chen, Hao and Poddar, Souradip and Zhao, Linran and Jia, Yaoyao and Pan, David Z},
	booktitle={Proceedings of the 2023 International Symposium on Physical Design},
	pages={84--92},
	year={2023}
}

@inproceedings{BBV+2024,
	title={Fast ml-driven analog circuit layout using reinforcement learning and steiner trees},
	author={Basso, Davide and Bortolussi, Luca and Videnovic-Misic, Mirjana and Habal, Husni},
	booktitle={2024 20th International Conference on Synthesis, Modeling, Analysis and Simulation Methods and Applications to Circuit Design (SMACD)},
	pages={1--4},
	year={2024},
	organization={IEEE}
}

@inproceedings{DBB+2025,
	title={Enhancing reinforcement learning for the floorplanning of analog ics with beam search},
	author={Della Rovere, Sandro Junior and Basso, Davide and Bortolussi, Luca and Videnovic-Misic, Mirjana and Habal, Husni},
	booktitle={2025 21st International Conference on Synthesis, Modeling, Analysis and Simulation Methods, and Applications to Circuits Design (SMACD)},
	pages={1--4},
	year={2025},
	organization={IEEE}
}

@INPROCEEDINGS{BBV+2025a,
	author={Basso, Davide and Bortolussi, Luca and Videnovic-Misic, Mirjana and Habal, Husni},
	booktitle={2025 Design, Automation \& Test in Europe Conference (DATE)}, 
	title={Effective Analog ICs Floorplanning with Relational Graph Neural Networks and Reinforcement Learning}, 
	year={2025},
	pages={1-7},
	doi={10.23919/DATE64628.2025.10992888}
}

@inproceedings{GCY2025,
	title={AnalogGenie: A Generative Engine for Automatic Discovery of Analog Circuit Topologies},
	author={Gao, Jian and Cao, Weidong and Yang, Junyi and Zhang, Xuan},
	booktitle={The Thirteenth International Conference on Learning Representations},
	year={2025},
	url={https://openreview.net/forum?id=jCPak79Kev}
}

@inproceedings{BGK+2023,
	title={Chip-chat: Challenges and opportunities in conversational hardware design},
	author={Blocklove, Jason and Garg, Siddharth and Karri, Ramesh and Pearce, Hammond},
	booktitle={2023 ACM/IEEE 5th Workshop on Machine Learning for CAD (MLCAD)},
	year={2023},
}

@article{CWR+2023,
	title={Chipgpt: How far are we from natural language hardware design},
	author={Chang, Kaiyan and Wang, Ying and Ren, Haimeng and Wang, Mengdi and Liang, Shengwen and Han, Yinhe and Li, Huawei and Li, Xiaowei},
	journal={arXiv preprint arXiv:2305.14019},
	year={2023}
}

@inproceedings{
    CPZ+2026,
    title={{OSIRIS}: Bridging Analog Circuit Design and Machine Learning with Scalable Dataset Generation},
    author={Giuseppe Chiari and Michele Piccoli and Davide Zoni},
    booktitle={The Fourteenth International Conference on Learning Representations},
    year={2026},
    url={https://openreview.net/forum?id=TIDaHgj0Yj}
}

@article{LFH+2021,
  title={Performance evaluation of deep CNN-based crack detection and localization techniques for concrete structures},
  author={Ali, Luqman and Alnajjar, Fady and Jassmi, Hamad Al and Gocho, Munkhjargal and Khan, Wasif and Serhani, M Adel},
  journal={Sensors},
  volume={21},
  number={5},
  pages={1688},
  year={2021},
  publisher={MDPI}
}

@article{CLL+2021,
  title={Deep neural network based vehicle and pedestrian detection for autonomous driving: A survey},
  author={Chen, Long and Lin, Shaobo and Lu, Xiankai and Cao, Dongpu and Wu, Hangbin and Guo, Chi and Liu, Chun and Wang, Fei-Yue},
  journal={IEEE Transactions on Intelligent Transportation Systems},
  volume={22},
  number={6},
  pages={3234--3246},
  year={2021},
  publisher={IEEE}
}

@INPROCEEDINGS{CGL+2024,
  author={Chiari, Giuseppe and Galli, Davide and Lattari, Francesco and Matteucci, Matteo and Zoni, Davide},
  booktitle={2024 Design, Automation \& Test in Europe Conference \& Exhibition (DATE)}, 
  title={A Deep- Learning Technique to Locate Cryptographic Operations in Side-Channel Traces}, 
  year={2024},
  pages={1-6},
  doi={10.23919/DATE58400.2024.10546758}
}

@INPROCEEDINGS{GCZ2024,
  author={Galli, Davide and Chiari, Giuseppe and Zoni, Davide},
  booktitle={2024 IEEE 42nd International Conference on Computer Design (ICCD)}, 
  title={Hound: Locating Cryptographic Primitives in Desynchronized Side-Channel Traces using Deep-Learning}, 
  year={2024},
  pages={114-121},
  doi={10.1109/ICCD63220.2024.00027}
}

@article{GCZ2025,
  title={Chameleon: A dataset for segmenting and attacking obfuscated power traces in side-channel analysis},
  author={Galli, Davide and Chiari, Giuseppe and Zoni, Davide},
  journal={IACR Transactions on Cryptographic Hardware and Embedded Systems},
  volume={2025},
  number={3},
  pages={389--412},
  year={2025}
}

\end{document}